%% file: emnlp2023.tex
\pdfoutput=1

\documentclass[11pt]{article}

\usepackage[]{EMNLP2023}

\usepackage{times}
\usepackage{latexsym}
\usepackage{comment}
\usepackage{arydshln}
\usepackage{enumitem}
\usepackage{booktabs}
\usepackage{graphicx}
\usepackage{cleveref}
\usepackage{natbib}

\usepackage[T1]{fontenc}

\usepackage[utf8]{inputenc}

\usepackage{microtype}
\usepackage{algorithm}
\usepackage{algpseudocode}

\usepackage{inconsolata}

\title{Paraphrase and Aggregate with Large Language Models for Minimizing Intent Classification Errors}

  \author{Vikas Yadav$^\dagger$, Zheng Tang$^\ddagger$, Vijay Srinivasan$^\ddagger$ \\
  ServiceNow$^\dagger$, Samsung Research America$^\ddagger$, USA \\
  {\tt vikas.yadav@servicenow.com, \{zheng.tang,v.srinivasan\}@samsung.com}}

\begin{document}
\maketitle
\begin{abstract}

  Large language models (LLM) have achieved remarkable success in natural language generation but lesser focus has been given to their applicability in decision making tasks such as classification. We show that LLMs like LLaMa can achieve high performance on large multi-class classification tasks but still make classification errors and worse, generate out-of-vocabulary class labels. To address these critical issues, we introduce Paraphrase and AGgregate (PAG)-LLM approach wherein an LLM generates multiple paraphrases of the input query (parallel queries), performs multi-class classification for the original query and each paraphrase, and at the end aggregate all the classification labels based on their confidence scores. We evaluate PAG-LLM on two large multi-class classication datasets: CLINC, and Banking and show 22.7\% and 15.1\% error reduction. We show that PAG-LLM is especially effective for hard examples where LLM is uncertain, and reduces the critical misclassification and hallucinated label generation errors. 

\end{abstract}

\input{emnlp2023-latex/Introduction}

\input{emnlp2023-latex/RelatedWork}

\input{emnlp2023-latex/Approach}

\input{emnlp2023-latex/Results}

\input{emnlp2023-latex/Analaysis}
\input{emnlp2023-latex/Limitations}

\bibliography{custom}
\bibliographystyle{emnlp2023}

\appendix

\end{document}

%% file: emnlp2023-latex/Introduction.tex
\section{Introduction}

Recent progress on generative AI has had a transformative effect on the field of NLP and ML \cite{brown2020language,llamapaper,chowdhery2022palm}. Large language models (LLM) have received more spotlight for generative tasks such as question answering, dialogue, summarization, etc \cite{peng2023instruction,open-llm-leaderboard}. We argue that key NLP tasks such as intent classification is widely utilized in real-world dialogue systems and thus should also be given high emphasis when evaluating LLMs, considering their proven capability to solve a wide range of NLP tasks \cite{open-llm-leaderboard}. 
In this work, we focus on studying LLMs for large intent classification tasks with two intent classification datasets: CLINC \cite{clinc} which has 150 classes and Banking \cite{casanueva2020efficient} which has 77 classes. Intent classification is vital in many real-world NLP systems, where mapping an input query to an in-domain class or rejecting it (if out-of-domain) significantly affects system performance \cite{shen2021enhancing}. We demonstrate that LLMs, such as LLaMa, achieve strong performance on both datasets but still make miscalssification and worse, new unseen label generation errors. Thus, our work focuses on addressing such critical errors in intent classification from LLMs.

\begin{figure*}[!h] 
\centering
\includegraphics[width=16cm]{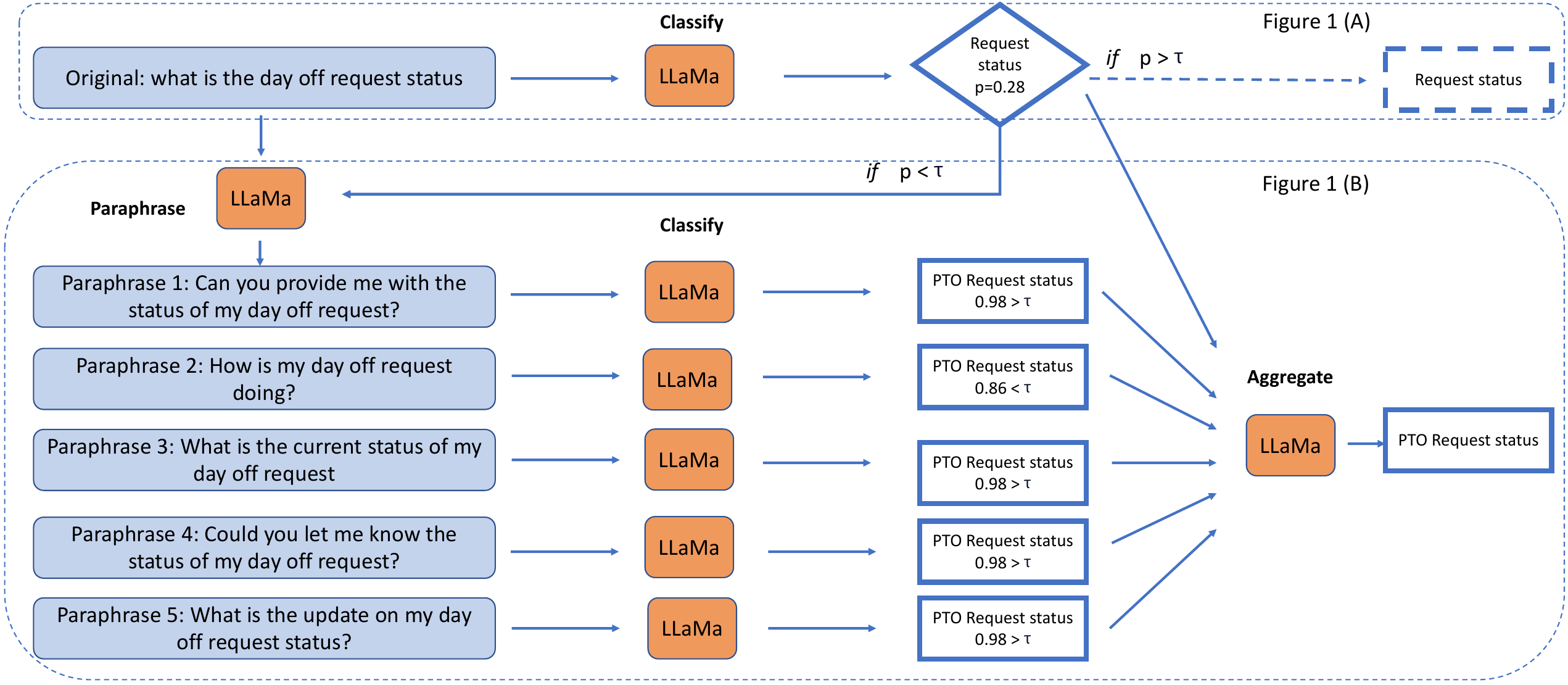} 
 \caption{\footnotesize Left figure depicting the flow process of PAG-LLM. 
On the left \cref{fig:examplefig}(A), LLM classifies the original query and only if the classification confidence is lower than $\tau$, 
 original query is given to the LLM for generating paraphrases which are then again given to the LLM for classification. Finally, LLM aggregates the predicted class labels from paraphrases and the original query. In the right figure table,  examples from CLINC are shown where LLM classifies incorrect label (top example) and out-of-vocabulory (OOV) class label (bottom example). In the top example, paraphrases generated by PAG-LLM enables correct classification decisions with high confidence scores. Thus, even simple majority voting aggregation leads to the correct class prediction. In the bottom example, only paraphrase2 from PAG-LLM enables correct classification while remaining paraphrases and the original query have OOV class labels. PAG-LLM aggregates texts of input, paraphrases, their labels and confidences to finally predict the correct class label.  }
  \label{fig:examplefig}
\end{figure*}

\begin{table*}[t]
   \centering
   \small
    \begin{tabular}{rlll}
     \hline
     & Input & Prediction & Confidence\\ \hline
     Original: & what is the day off request status & request\_status & \textcolor{red}{0.28}\\
     Paraphrase 1:& Can you provide me with the status of my day off request? & pto\_request\_status & \textcolor{teal}{0.98}\\
     Paraphrase 2:& How is my day off request doing? & pto\_request\_status & \textcolor{red}{0.86}\\
     Paraphrase 3:& What is the current status of my day off request? & pto\_request\_status & \textcolor{teal}{0.98}\\
     Paraphrase 4:& Could you let me know the status of my day off request? & pto\_request\_status & \textcolor{teal}{0.98}\\
     Paraphrase 5:& What is the update on my day off request status? & pto\_request\_status & \textcolor{teal}{0.98}\\
     \hline
     \multicolumn{2}{r}{Final Prediction: pto\_request\_status {Aggregation EXAMPLE}}&
     \multicolumn{2}{r}{Gold Label: pto\_request\_status} \\
     \hline
     Original: & what is the reason humans even exist & explain\_meaning\_of\_life & \textcolor{red}{0.11}\\
     Paraphrase 1:& What is the purpose of human existence? & meaning\_of\_life & \textcolor{teal}{0.32}\\
     Paraphrase 2:& Why do humans exist in the world? & The meaning of life & \textcolor{red}{0.05}\\
     Paraphrase 3:& What is the cause of human existence? & $NULL$ & \textcolor{red}{0.07}\\
     Paraphrase 4:& What is the explanation for the existence of humans? & explain\_life & \textcolor{red}{0.21}\\
     Paraphrase 5:& What is the rationale behind the existence of human beings? & rational\_existence & \textcolor{red}{0.08}\\
     \hline
     \multicolumn{2}{r}{Final Prediction: meaning\_of\_life \textcolor{red}{Aggregation EXAMPLE}} &
     \multicolumn{2}{r}{Gold Label: meaning\_of\_life} \\
     \hline
   \end{tabular}
   \label{Exampletable}
   \caption{\small Examples from CLINC, where LLM classifies incorrect label (1st example) and out-of-vocabulory (OOV) class label (2nd example). In the 1st example, paraphrases generated by PAG-LLM enables correct classification decisions with high confidence scores. Thus, even simple majority voting aggregation leads to the correct class prediction. In the 2nd example, only paraphrase2 from PAG-LLM enables correct classification while remaining paraphrases and the original query have OOV class labels. PAG-LLM aggregates texts of input, paraphrases, their labels and confidences to finally predict the correct class label.}
\end{table*}


In recent approaches like self-consistency \cite{selfcons}, outputs of LLM from multiple reruns on the same input (with different temperatures or sampling techniques) can be aggregated for reducing LLM errors. In \cref{ID_result_table}, we show that such approaches are less applicable to straightforward classification tasks where the generated text is a single word or phrase representing the class label. In short response generations like single word for class label, tokens with highest generation probability should be selected as the final class label text.   Similarly, while sampling techniques or temperature scaling \cite{vijayakumar2016diverse, holtzmancurious} are vital for promoting diversity, they may be lesser applicable for fixing errors in classification tasks, where selecting tokens with the highest probabilities is necessary to predict the class label. Hence, as an alternative solution, we propose a (p)araphrasing and (ag)gregating approach (PAG) to fix LLM errors on intent classification task, where input query is paraphrased to perform intent classification on its multiple variations. Our approach is inspired by observations that often user queries are unclear which when rephrased, improve downstream systems \cite{brabant2022coqar}. PAG-LLM leverages the versatility of LLMs to perform three tasks: paraphrasing, intent classification, and aggregation. We first generate N paraphrases of the input query, then generate classification predictions for the original query and its N paraphrases and finally, aggregate the classification decisions for generating the final prediction, all using LLMs.  
Our key findings are as follows:

\begin{itemize}[leftmargin=*]
    \item We study LLMs on large intent classification task and show substantial error reduction through our proposed PAG-LLM approach. PAG-LLM reduces error by 22.7\% on CLINC and 15.1\% on Banking intent classification datasets. PAG-LLM also shows improvements in the out-of-domain intent classification setting \cite{zhou2022knn} with 3.2\% and 1.5\% absolute F1 score improvement in CLINC and Banking respectively. 
    \item  We show that PAG-LLM can be selectively applied to low-confidence classification cases to potentially lower the inference cost (\cref{Analysis}). We also present analyses showing distribution of misclassification and hallucinated label generation error corrections with PAG-LLM.  
\end{itemize}

%% file: emnlp2023-latex/RelatedWork.tex
\section{Related Work}




%

A few recent works have evaluated LLMs on classification tasks such as commonsenseQA, multiple-choice QA, and Boolean QA that have context documents and small number of classes \cite{llamapaper, zhang2022opt, clark2019boolq} but lesser focus has been given on understanding and fixing their errors. Some of the recent approaches like self-consistency \cite{selfcons} and chain-of-thought (COT) \cite{cot} have shown to improve performance but mainly for arithmetic and symbolic reasoning tasks such as GSM8K, ArithmeticQA, StrategyQA \cite{cobbe2021training}. 
As discussed previously, these approaches are intuitively not promising for (context-free) intent classification tasks because of single word (or phrase) generation for predicting class label \cite{pitis2023boosted}. 
Approaches like our PAG-LLM, where an input query is asked in multiple different ways via paraphrasing can be an effective solution in a wide variety of classification tasks. For example, PAG-LLM via paraphrasing is able to solve both misclassification error and OOV label generation error as shown in right \cref{fig:examplefig} example table. 

Paraphrasing is a proven effective solution for evaluating and enhancing the robustness of NLP systems \cite{pararobust}. Previously, paraphrases were generated using a separate encoder-decoder module and then provided to a different neural end-task model, resulting in lower end performance. \cite{pararobust, falke2020leveraging, cho2019paraphrase}. Contrary to these, PAG-LLM leverages LLMs' versatility to handle all paraphrasing, classification, and aggregation, showcasing its benefits through significant error reduction and performance improvements of the LLM classifier. Some of the recent work also utilize in-context few shot prompting for intent classification \cite{milios2023context}. We compare to these previous works in \cref{ID_result_table}. 

%% file: emnlp2023-latex/Approach.tex
\section{Approach}

We focus and limit our experiments to only open LLMs\footnote{Our organization prohibts usage of licensed and human feedback learning LLMs such as ChatGPT for privacy reasons. We have included results of PAG-LLM with ChatGPT in appendix.} such as LLaMa \cite{llamapaper}. PAG-LLM process flow is formulated in \Cref{alg:PAG-llm} and a walkthrough example is shown in left \cref{fig:examplefig}. The classifier LLM\textsubscript{classify} classifies the input query $Q_i$ to class $C_i$ with $p_i$ confidence. If the classification confidence $p_i$ > $\tau$, $C_i$ is considered as the final label (depicted in left \cref{fig:examplefig}(A)) else LLM\textsubscript{paraphrase} generates n paraphrases $[PQ_i^1 ... PQ_i^n]$ of $Q_i$. Then, LLM\textsubscript{classify} classifies each of the paraphrased queries predicting classes $Cp_i$. LLM\textsubscript{aggregate} then aggregates original and its paraphrased queries, their predicted classes along with their confidences to predict the final class $C_i$. This is depicted in left \cref{fig:examplefig}(B)).  


\begin{algorithm}
\caption{PAG-LLM algorithm}\label{alg:PAG-llm}
\begin{algorithmic}
\State $Q_i \gets $ Input query
\State $C_i, p_i \gets LLM_{classify}(Q_i)$
\If{$p_i > \tau $}
   \State return $C_i$ \Comment{{\tiny \Cref{fig:examplefig}(A)}}
\Else \Comment{{\tiny If confidence is $< \tau$, invoke PAG-LLM}}
    \State $[PQ_i^1 ... PQ_i^n] \gets LLM_{paraphrase}(Q_i)$
    \State Predictions $= [C_i, p_i, Q_i] $
\For{$PQ \gets PQ_i^1$ to $PQ_i^n$}                    
        \State {$Cp_i, pp_i \gets LLM_{classify}(PQ)$}
        \State Predictions.insert($[Cp_i, pp_i, PQ]$)
    \EndFor
\State $C_i, p_i \gets LLM_{Aggregate}($Predictions$)$  
\State return $C_i$
\EndIf

\end{algorithmic}
\end{algorithm}



As shown in \Cref{fig:examplefig}, PAG uses LLM for paraphrasing, classification, and aggregating. We supervised finetune (SFT) LLaMa for each of these 3 tasks until convergence. 


\noindent{\bf Classification:} LLM\textsubscript{classify} is supervised finetuned (SFT) on training data of CLINC and Banking-50 for the classification task. 

\noindent{\bf Paraphrasing:}  For paraphrasing, we finetune LLaMa on 419K ChatGPT paraphrasing dataset \footnote{ChatGPT paraphrase dataset - \url{https://huggingface.co/datasets/humarin/chatgpt-paraphrases}} which has 5 paraphrases of each query from Quora dataset, texts from SQuAD2.0~\cite{rajpurkar2018know} and CNN datasets~\cite{chen2016thorough}. LLM\textsubscript{paraphrase} generates 5 paraphrases\footnote{We limited our experiments to 5 paraphrases because our training data - ChatGPT paraphrasing dataset has only 5 paraphrase outputs for each query.} of the original query (as shown in left \cref{fig:examplefig}(B)) which are all fed to the LLM\textsubscript{classify} individually for classification.


\noindent{\bf Aggregating:} In LLM\textsubscript{aggregate} finetuning for aggregation, we first generate predictions on original and paraphrased queries on the validation data. We simply concatenate the original query, all paraphrased queries, their prediction labels (along with their corresponding confidence scores) as input to train LLaMa to predict the final output label.


\subsection{Need for finetuning}
Intent classification is a critical first task that impacts downstream performance of real-world dialogue systems. Thus, having high accuracy and maximum plausible control for preserving privacy (for e.g., in healthcare, and legal domains) is critical which can be achieved by full SFT of open LLMs \cite{wang2023aligning, zhang2023instruction}. A few shortcomings of open instruction tuned LLMs that necessitate SFT are as follows:  
\begin{itemize}[noitemsep,leftmargin=*]
    \item Parsing generated output: OpenLLMs with zero or few-shot prompts often deviate from instructions, generating relevant but off-target responses \cite{jin2022good}. We observed similar issues with open instruction finetuned (IFT) LLMs such as Vicuna and Alpaca where generated response had differnt types of text in addition to the intent class label.
    \item Intent label informativeness - We observed that intent label text were often confusing and overlapped with other intent classes. Showing examples of how queries are mapped to each intent class may not be feasible for LLM prompting input length in a large-intent classification task (>150 classes). In-context learning (ICL) is an effective solution but LLM performance is greatly dependent on retrieval accuracy \cite{milios2023context}. We show comparison to ICL in \cref{ID_result_table}.
    \item Prompt sensitivity - Prompt selection has shown to significantly  affect openLLMs performances \cite{pearce2023examining}. To avoid such variations, we fine-tune LLMs in all our experiments until convergence on validation set. 
\end{itemize}

\subsection{Datasets}
We present our results on two large multi-class classification datasets: CLINC (with 150 intent classes) and Banking-50 (with 77 intent classes). In the Banking-50 dataset, 50\% of the intent labels are masked and labeled as OOD (out-of-domain), while the remaining 50\% (i.e., 38 labels) are retained as original. Following previous work \cite{zhou2022knn}, we train the LLM exclusively on the in-domain training data (2.1k samples for Banking-50 and 15k for CLINC). We showcase PAG evaluations in two settings:

\begin{itemize}[noitemsep,leftmargin=*]
\item{In-domain:}(ID) is the straightforward multiclass classification task where input query is labeled into one of the predefined classes. Here, we only evaluate on in-domain test inputs (4.5K of CLINC and 1.5K of banking). Results are shown in \cref{ID_result_table}.
\item {With Out-of-domain:}(ID+OOD) - We show evaluations on the full test dataset (5.5K queries of CLINC and 3.1K of banking) with both ID+OOD queries in \cref{OD_result_table}. In this setting, a query is labelled as OOD if it does not belong to the predefined ID classes. We consider a prediction to be OOD if the classification confidence is lower than a decision threshold or the generated label is out of label vocabulary list. Predictions above the threshold are mapped to the corresponding ID class. Techniques like PAG-LLM can be crucial for inputs that are closer to such important decision threshold. 
\end{itemize}

\begin{table*}
 \centering
\footnotesize
\begin{tabular}{llll|cc|cc}
\toprule
\# & Exp & Aggregate & Num & \multicolumn{2} {c|}{CLINC} & \multicolumn{2} {c}{Banking-50} \\
\cline{5-6} \cline{7-8} 
& & method & runs  & ID F1 & Error Reduct. & ID F1 & Error Reduct.\\
\midrule
P1 & ICL (5 shot) LLaMa-7B \cite{milios2023context} & - & - &  88.58 & & 84.42 & \\
P2 & ICL (10 shot) LLaMa-7B \cite{milios2023context} & - & - &  91.73 & & 87.63 & \\
P3 & ICL (5 shot) OPT-13B \cite{milios2023context} & - & - &  85.27 & & 81.23 & \\
P4 & ICL (10 shot) OPT-13B \cite{milios2023context} & - & - &  89.24 & & 85.65 & \\
\hdashline
1 & SFT LLaMa-7B & - & 1x & 96.29 & - & 94.04 & -  \\
\hdashline
2 & Self-consistency (top\_k)\cite{selfcons} & Vote & 6x & 96.15 & \textcolor{red}{+3.8} & 92.89 &  \textcolor{red}{+19.3} \\
3 & Self-consistency (Temp.)\cite{selfcons} & Vote & 6x & 96.18 & \textcolor{red}{+3.0} & 93.62 & \textcolor{red}{+7.0} \\
\hdashline
4&   SFT LLaMa-7B (Rand-Seed) & Vote & 6x & 96.51 & \textcolor{teal}{-5.9} & 94.41 & \textcolor{teal}{-6.2} \\ 
5&   PAG (All queries) & Vote & 6x & 96.28 & \textcolor{red}{+0.2} & 93.52 & \textcolor{red}{+8.6}\\
6&   PAG (low conf < $\tau$) & Vote & 2.6x & 96.32 & \textcolor{teal}{-0.8} & 94.42 & \textcolor{teal}{-6.4} \\
   \hdashline
7& PAG (All queries)   & LLM & 6x & 97.05 & \textcolor{teal}{-20.4} & {94.85} & \textcolor{teal}{13.6} \\
8&PAG (low conf < $\tau$)  & LLM & 2.6x &  {\bf 97.13} & \textcolor{teal}{\bf -22.7} & {\bf 94.94} & \textcolor{teal}{{\bf -15.1}} \\

\bottomrule
\end{tabular}
\caption{\label{ID_result_table} \footnotesize Performance on in-domain (ID) CLINC and Banking(50\%) datasets. We use our same SFT LLaMa-7B from row 1 in self-consistency runs also (row2 and row3). Rand-seed denotes ensembling of 6 different SFT LLaMa trained with different random seeds. Vote denotes majority voting strategy to select the final label. 
For row 6 and 8, we tune the confidence threshold (denoted by $\tau$=0.98, 0.90 for CLINC and Banking) on dev data and only the queries with lower classification confidences are given to PAG-LLM. 
Best numbers are shown in bold. P1, P2, P3, P4 represent 4 in-context learning (ICL) baselines from previous work.}
\end{table*}

\begin{table*}[t]
\centering
\begin{tabular}{ll|cccc|cccc}
\toprule
Exp & Agg. & \multicolumn{4} {c|}{CLINC} & \multicolumn{4} {c}{Banking-50} \\
\cline{3-6} \cline{7-10} 
&Meth. & ID & OOD  &All  &Avg & ID  & OOD  &All  &Avg \\
& & F1 & F1 & F1 & (ID+OOD) & F1 & F1 & F1 &(ID+OOD) \\

\midrule
SFT LLaMa-7B & - & 88.5 & 88.85 & 88.51  & 88.67 & 77.04 & 77.05 & 77.49 & 77.26 \\
\hdashline
 PAG (All queries) & Vote & 91.51 & 92.56 & 91.51 & 92.03 & 77.14 & 77.49 & 77.15 & 77.31 \\
   PAG (conf < $\tau$) & Vote & 91.67 & {\bf 92.58} & 91.68 &  92.13 & 78.28 & 77.74 & 78.28 & 78.02\\
   \hdashline
PAG (conf < $\tau$) & LLM  & {\bf 92.04} &  92.54 & {\bf 92.05}  & {\bf 92.29} & {\bf 78.52} & {\bf 77.8} & {\bf 78.5} & {\bf 78.16} \\
\bottomrule
\end{tabular}
\caption{\label{OD_result_table} \footnotesize Performances on overall test datasets (ID and OOD inputs) of CLINC and Banking(50\%). Notations are same as in \cref{ID_result_table}. All F1 denotes the macro F1 score for all ID + 1 OOD classes (i.e., 151 classes for CLINC and 38 for banking). Avg denotes average of ID and OOD F1 scores.}
\end{table*}

\subsection{Hyperparameters}
We finetuned LLaMa-7B on 4 A100 GPUs using huggingface library \cite{wolf2019transformers} for 4 epochs, learning rate=1e-5, batchsize=4, and gradient accumulation step=4 on Banking-50 and CLINC training sets. For paraphrasing and aggregation, we SFT LLaMa-7B for 3 epochs.

%% file: emnlp2023-latex/Results.tex
\section{Results}

We implemented three baselines with SFT LLaMa-7B based on self-consistency (SC) \cite{selfcons} (with top\_k sampling and temperature) and training N different LLaMa-7B classifiers with different random seed ("Rand-seed" in \cref{ID_result_table}). We also show comparison with in-context learning based baselines from previous works in \cref{ID_result_table}. Our PAG-LLM method with LLaMa-7B are shown in row 5-8. Row 5 and 6 use simple voting for aggregation of the classification from original query and its 5 paraphrases i.e., LLM is utilized only for paraphrasing and intent classification while voting is used for aggregation. PAG-LLM in Row 7 and 8 utilizes LLM for aggregation also. In row 5 and 7, paraphrasing is done for all the input queries whereas in row 6 and 8, paraphrasing+aggregation is done only on low confidence input queries i.e., where LLM intent classification is uncertain. Results of in-domain evaluations are shown in \cref{ID_result_table} and overall (ID+OOD) evaluations in \cref{OD_result_table}.

\begin{enumerate}[label={\bf(\arabic*)},itemsep=0em,topsep=0em,  wide, labelwidth=!, labelindent=0pt] 
\item {\bf ID-evaluations}: SFT LLaMa-7B achieves high performance in ID classification, outperforming previous ICL based LLM classifiers by subtantial margins (row 1 VS P1-P4 rows). Performance of SFT LLaMa-7B is further improved by our full PAG approach as shown in row 7 and 8. Specifically, as shown in row 8 of \cref{ID_result_table}, PAG-LLM results in 22.7\% and 15.1\% error reduction in CLINC and Banking-50 datasets respectively. Aggregation using LLM is always better than voting emphasizing its usefulness within PAG (row 7,8 vs 5,6).  

\item {\bf Baseline comparisons}: As expected, we observed slightly lower performance from aggregating predictions from top\_k sampling (row 2) and temperature=0.9 (row 3) confirming self consistency like approaches may not work in such large (context-free) intent classification tasks. Our PAG-
Aggregating predictions from 6 different LLM classifiers (trained with different random seeds) is slightly better than voting aggregation with PAG (row4 vs row6) but arguably having N different LLM classifiers is not practical. On the other hand, PAG-LLM can SFT a single LLM to paraphrase, classify and aggregate all predictions. 

\item {\bf Aggregating Lower Confidence:} Paraphrasing and aggregating only lower confidence queries (row 6,8) results in higher performance compared to running PAG-LLM on all of the inference queries (row 5,7). This emphasizes the practical potential of PAG-LLM as paraphrasing and aggregation is needed only on 32\% of the test inputs that have confidence below 98\% in CLINC (thus resulting in 0.32*5 (paraphrased queries) + 1 (original query) = 2.6x number of inference runs). 
\item {\bf OOD-evaluations}: Aggregation is effective for distinguishing ID from out-of-domain (OOD) queries \cite{zhou2022knn}, after which it can be classified into one of the ID classes. In \cref{OD_result_table}, PAG-LLM shows nearly 3.9\% F1 improvement in OOD classification while also improving the ID F1 by 3.5\% on CLINC. The improvements in Banking-50 are relatively smaller (1.5\% F1) possibly due to its longer class labels compared to CLINC.


\end{enumerate}

%% file: emnlp2023-latex/Analaysis.tex
\section{Analysis}
\label{Analysis}

\begin{figure}
\includegraphics[width=1\columnwidth]{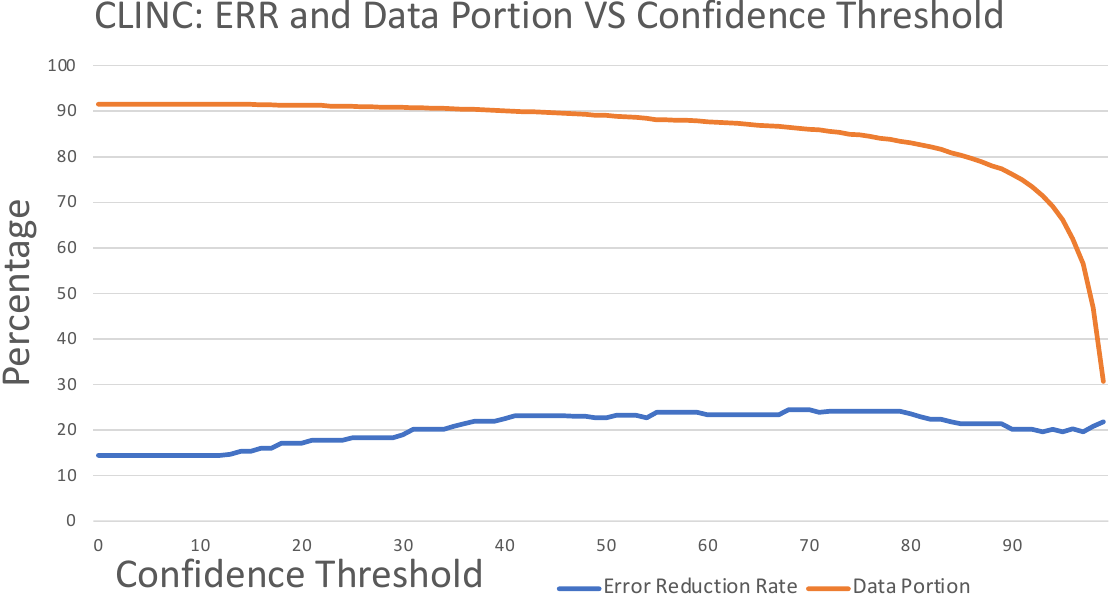} 
 \caption{\footnotesize Plot showing portion of inference data and error rate reduction on CLINC with increasing classification threshold ($\tau$). }
  \label{fig:ERR_plot}
\end{figure}

Our work highlights high performance of SFT LLMs like LLaMa-7B for intent classification. PAG-LLM further improves the performance by substantially reducing errors of SFT LLaMa-7B model, especially for uncertain input queries with low classification confidence (row 8, \cref{ID_result_table}). We analyzed error rate reductions (ERR) and test data portion with varying confidence threshold ($\tau$). As shown in \cref{fig:ERR_plot}, ERR remains similar but data proportion drops substantially with higher $\tau$ (i.e., $\tau>0.7$). With the validation data tuned $\tau=98\%$, only 32\% of the queries need to be paraphrased with PAG-LLM achieving 22.7\% error reduction in CLINC, (row8 of \cref{ID_result_table}). We further analyzed the correction distribution of PAG-LLM. Out of 22.7\% error correction on CLINC, 14.4\% were for out-of-vocabulory (OOV) class label generation and remaining 8.3\% for misclassification errors (shown in right \cref{fig:examplefig} example table). Similarly, out of 15.1\%, 8.1\% and 7.0\% error reduction were for OOV and misclassification errors in Banking dataset.

Overall, our work presents a new method to reduce errors of LLM for intent classification. PAG-LLM approach utilizes LLaMa-7B for each of the three components for intent classification, paraphrasing and aggregation leading to 22.7\% and 15.1\% error reduction on CLINC and Banking respectively. We show that PAG-LLM is effective in error reduction even for high performing tasks such as intent classification. Such findings are encouraging for future works to apply PAG-LLM to other key information retrieval and NLP tasks. 


%% file: emnlp2023-latex/Limitations.tex
\section{Limitations}


In this short paper, we have focused on presenting our novel PAG-LLM approach for error reduction in intent classification. We have summarized a few limitations of our study below:
\begin{itemize}[noitemsep,leftmargin=*]
    \item Effect of paraphrasing - We also experimented with randomly selected 210K paraphrasing data out of the 419K ChatGPT paraphrasing dataset. We did not observe any substantial changes in our result findings but in-depth exploration of paraphrasing quality as future work could further enhance PAG-LLM approach.    
    \item Extension to other LLMs - As SFT is a compute expensive process, we showcase our experiments with LLaMa-7B. PAG-LLM can be applied to any other open LLM and we will provide our github codes for easily extending to any LLM available in Huggingface model repositories.
\end{itemize}